\documentclass[conference]{IEEEtran}
\IEEEoverridecommandlockouts
\makeatother
\usepackage{cite}
\usepackage{amsmath,amssymb,amsfonts}
\usepackage{algorithmic}
\usepackage{graphicx}
\usepackage{textcomp}
\usepackage{xcolor}
\usepackage{amsmath}

\usepackage{romannum}
\usepackage{bbold}
\usepackage{amssymb}
\usepackage[linesnumbered,ruled,vlined]{algorithm2e}
\usepackage{array}
\usepackage{booktabs}
\usepackage{enumitem}
\usepackage{algorithmic}
\usepackage{setspace}
\usepackage{hyperref}
\usepackage{subfig}

\def\BibTeX{{\rm B\kern-.05em{\sc i\kern-.025em b}\kern-.08em
    T\kern-.1667em\lower.7ex\hbox{E}\kern-.125emX}}

\makeatletter
\newcommand{\linebreakand}{%
  \end{@IEEEauthorhalign}
  \hfill\mbox{}\par
  \mbox{}\hfill\begin{@IEEEauthorhalign}
}
\makeatother

\begin{document}

\title{Yet Meta Learning Can Adapt Fast,\\ It Can Also Break Easily}

\author{\IEEEauthorblockN{Han Xu}
\IEEEauthorblockA{\textit{Dept. of Comp. Sci. and Engr.} \\
\textit{Michigan State University}\\
East Lansing, MI \\
xuhan1@msu.edu}
\and
\IEEEauthorblockN{Yaxin Li}
\IEEEauthorblockA{\textit{Dept. of Comp. Sci. and Engr.} \\
\textit{Michigan State University}\\
East Lansing, MI \\
liyaxin1@msu.edu}
\and
\IEEEauthorblockN{Xiaorui Liu}
\IEEEauthorblockA{\textit{Dept. of Comp. Sci. and Engr.} \\
\textit{Michigan State University}\\
East Lansing, MI \\
xiaorui@msu.edu}
\linebreakand
\IEEEauthorblockN{Hui Liu}
\IEEEauthorblockA{\textit{Dept. of Comp. Sci. and Engr.} \\
\textit{Michigan State University}\\
East Lansing, MI \\
liuhui7@msu.edu}
\and
\IEEEauthorblockN{Jiliang Tang}
\IEEEauthorblockA{\textit{Dept. of Comp. Sci. and Engr.} \\
\textit{Michigan State University}\\
East Lansing, MI \\
tangjili@msu.edu}
}

\maketitle

\begin{abstract}
 Meta learning algorithms have been widely applied in many tasks for efficient learning, such as few-shot image classification and fast reinforcement learning. During meta training, the meta learner develops a common learning strategy, or experience, from a variety of learning tasks. Therefore, during meta test, the meta learner can use the learned strategy to quickly adapt to new tasks even with a few training samples. However, there is still a dark side about meta learning in terms of reliability and robustness. In particular, 
  {\it is meta learning vulnerable to adversarial attacks?} In other words, would a well-trained meta learner utilize its learned experience to build wrong or likely useless knowledge, if an adversary unnoticeably manipulates the given training set? Without the understanding of this problem, it is extremely risky to apply meta learning in safety-critical applications. Thus, in this paper, we perform the initial study about adversarial attacks on meta learning under the few-shot classification problem. In particular, we formally define key elements of adversarial attacks unique to meta learning and propose the first attacking algorithm against meta learning under various settings. We evaluate the effectiveness of the proposed attacking strategy as well as the robustness of several representative meta learning algorithms. Experimental results demonstrate that the proposed attacking strategy can easily break the meta learner and meta learning is vulnerable to adversarial attacks. The implementation of the proposed framework will be released upon the acceptance of this paper.
  \\
\end{abstract}


\section{Introduction}

Deep neural networks have achieved extraordinary accomplishments in numerous domains such as computer vision~\cite{devlin2018bert} and natural language processing~\cite{krizhevsky2012imagenet}. To achieve satisfying performance, they usually require large-scale training data. Therefore, learning efficiently and effectively with small data has become one desirable property for modern machine learning techniques. 
Meta learning (or learning to learn) algorithms~\cite{finn2017model, ravi2016optimization} have been widely used to improve the efficiency of learning new tasks. 
Meta learning models are often composed of two nested parts, a \textit{meta learner} and an \textit{adapted learner}. Generally speaking, a \textit{meta learner} is trained on a lot of learning tasks to build a common learning strategy to solve these tasks. 
In each task, the \textit{meta learner} produces an \textit{adapted learner} 
which works at the level of specific task such as an image classifier or an object detector. 
During the meta-test phase, a meta learner generalizes its learning strategy to quickly tackle new tasks, only based on a few teaching episodes.

The success of meta learning algorithms has encouraged their applications in many safety-critical tasks, including face identification~\cite{zhou2018deep, guo2020learning}, object detection~\cite{wang2019meta, schwartz2018delta} and robotics imitation learning~\cite{finn2017one}. However, the reliability and robustness issues of meta learning algorithms have seldom been investigated and evaluated, which exposes the applications of meta learning to highly potential risk, especially with the existence of adversarial attackers. The concept of adversarial attack or adversarial example was originally proposed by \cite{szegedy2013intriguing} and \cite{goodfellow2014explaining}, which focus on deep neural networks (DNNs) for image classification problems. These adversarial examples are manually crafted images which have imperceptible difference with clean images but can mislead the DNN models to give totally wrong prediction. Similar phenomena were also found in other data domains, including graph domain~\cite{jin2020adversarial, jin2020graph, dai2018adversarial}, and language processing~\cite{liu2020chat}. The risk and concern of applying DNNs models in safety-critical missions under adversarial attacks have been revealed and highlighted such that more efforts have been paid on improving their robustness. However, the robustness of meta learning approaches against adversarial attack is still an open question and there is a pressing need to bridge this gap.

\begin{figure}
\centering
  \includegraphics[width=1\linewidth]{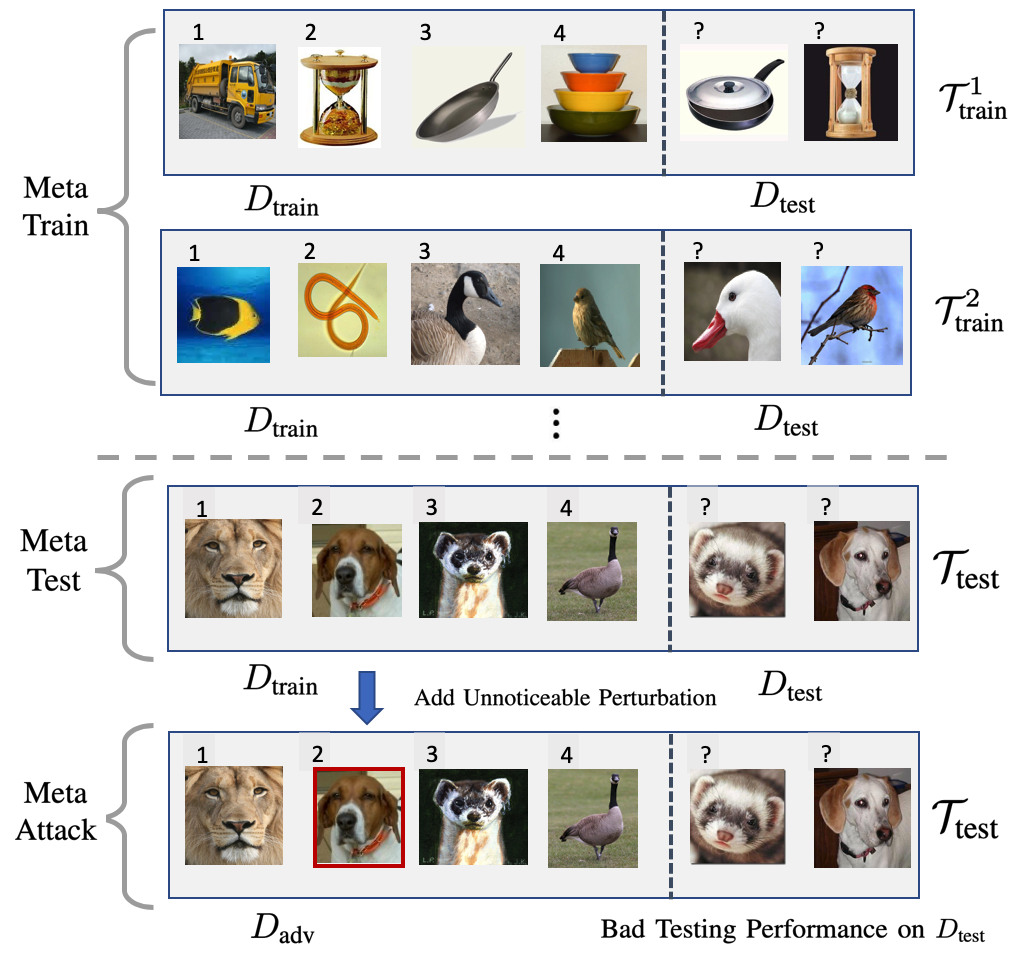}
  \caption{For few-shot classification tasks, during the meta test phase, an adversary can insert unnoticeable perturbation to one training sample of $D_{\text{train}}$ in the meta test task $\mathcal{T}^i_{\text{test}}$, causing the adapted model which is trained on $D_{\text{adv}}$ with much lower accuracy. }
  \label{fig:main}
\end{figure}

In this work, we are devoted to studying the adversarial robustness issues of meta learning with the focus on their application in few-shot classification problems. Specifically, we concentrate on attacking the meta learners, instead of the adapted learners. Since a meta learner acts as a function to output a good classification model based on a small teaching dataset, we ask the question: \textit{if an attacker slightly manipulates the teaching data of a meta learner, does the meta learner still produce a reliable classifier?} Compared to traditional attacks, the meta learning attacks face unique challenges. First, traditional attacks aim to mislead the machine learning model itself to give wrong prediction on the adversarially perturbed samples. While, meta attacks proposed in this work target on letting the meta learner produce a ``malicious'' machine learning model. This malicious model might have overall bad generalization to any unseen test samples or misclassify some specific items to a different class even when the test samples are well protected and not adversarial. In essence, the meta attacks can have much more severe consequences than traditional attacks on individual machine learning models. Second, to achieve attacking on unseen test samples is a challenging goal since it will require the generalization ability of the attacking strategy that could further uncover the vulnerability and unreliability of meta learning. Third, the concept of ``unnoticeable'' perturbation in meta learning needs to be redefined because the input of the meta learner is usually a sample set instead of an individual sample. For instance, in the case of few-shot image classification problems as showed in Fig.~\ref{fig:main}, the input of a meta learner is a set of images. In this case, in addition to guaranteeing small sample perturbation, we should also manage to attack the meta learner by fewest number of perturbed samples. Last but not least, the meta learning models might have intrinsically different architectures compared with DNNs models. Popular meta learning models include (1) optimization-based methods like MAML~\cite{finn2017model}; (2) model-based methods such as SNAIL~\cite{mishra2017simple} and (3)  metric-based models like Prototypical Net~\cite{snell2017prototypical}. These models have very heterogeneous designs which brings in further difficulty for a systematic evaluation on their robustness.


With the establishment of these differences and  challenges for attacking meta learning algorithms, we have demonstrated that dedicated efforts are desired to investigate the robustness of meta learning algorithms. In this work, we make an attempt for such investigation and our key contributions can be summarized as follows:

\begin{itemize}
    \item For the first time, we formally define the key elements for attacking meta learning algorithms, including adversarial goal and unnoticeable perturbation;
    \item We provide a new formulation of the objective function for meta attack with both targeted and non-targeted attacks under newly defined perturbation constraint;
    \item 
    A novel meta attack algorithm MetaAttacker is proposed to optimize the proposed objective function such that adversarial input set can be efficiently computed for diverse and complicated victim model structures;
    \item We systematically evaluate the reliability and robustness of meta learning through different meta learning frameworks including MAML, SNAIL and Prototypical. 
    Extensive experiments show that the proposed attacking algorithm can
    easily break the meta learner. 
    It suggests that meta learning approaches are vulnerable to the proposed meta attack, 
    which reveals its risk in safety-critical applications.
\end{itemize}


\noindent \textbf{Outline of the Paper} The rest of this paper is organized as follows. We briefly review related works about meta learning and adversarial attacks in Section~\ref{rel}. In Section~\ref{meta}, we introduce the basics and notations in meta learning. In Section~\ref{threat} we define the setup of the threat model while presenting the meta attack algorithm in Section~\ref{sec:alg}. Experimental results are presented in Section~\ref{exp}.  Finally we conclude our work with future work in Section~\ref{sec:conclusion}. 




\section{Related Work}\label{rel}
In line with the focus of this work, we briefly introduce the related works about meta learning and adversarial attacks.

\subsection{Meta Learning}


There exist various ideas to construct the meta learning models including Optimization-Based Meta Learner, Model-Based Meta Learner and Metric-Based Meta Learner. 

\noindent\textbf{Optimization-based meta learners} imitate the optimization process of training a machine learning model. For instance, Model-Agnostic Meta Learning (MAML)~\cite{finn2017model} learns a set of hyperparameters for a gradient descent process. When facing a new task, it can quickly adapt to a new classification model by running gradient descent in only several steps.

\noindent\textbf{Model-based meta learning models}~\cite{santoro2016one, mishra2017simple} 
directly take form as a neural network, usually a sequence model like LSTM~\cite{hochreiter1997long}. When facing a new task, the training samples act as parameters of the neural network, and outputs the predictions for test samples.

\noindent\textbf{Metric-based meta learners} learn the strategies to compare whether two samples are from the same class. Typically, this type of meta learner works with a DNN model which learns the embeddings of both training and test images. Then, they compare whether two images embeddings belong to the same class, based on different metrics, such as similarity functions~\cite{koch2015siamese, vinyals2016matching, snell2017prototypical}, SVM classifiers~\cite{lee2019meta} and ridge regression model~\cite{bertinetto2018meta}. 


In this work, we choose one representative algorithm in each category to explore our robustness characteristics, including MAML~\cite{finn2017model}, SNAIL~\cite{mishra2017simple} and Prototypical Networks~\cite{snell2017prototypical}, with the hope to have an overview of the robustness performance for different types of meta learners. 

\subsection{Adversarial Attacks}
The notion of adversarial attacks or adversarial examples of deep learning models was first introduced by \cite{szegedy2013intriguing, goodfellow2014explaining} on the image domain. Specifically, for a well-trained image classifier, an adversary can almost always generate unnoticeable perturbations on given images, and fool the classifier to make wrong predictions. The existence of adversarial examples demonstrates that deep learning models have non-robust properties, and could be unreliable when applied on safety-critical tasks. At the same time, adversarial attacks are also founded in other domains, such as graph structured data~\cite{ma2019attacking, xu2019adversarial} and text data~\cite{liu2019say, ebrahimi2017hotflip}. Typically, due to different adversarial goals, adversary's capacity and attacking phase, different adversarial attacks methods can be categorized as targeted and non-targeted attacks, and evasion and poisoning attacks. In targeted attack, the adversary aims to induce the classifier to give a specific label to the perturbed test sample while in non-targeted attack, the adversary only wants the classifier to predict incorrectly without specifying a label.  In evasion attacks, the model are fixed and usually have good performance on benign testing samples. The adversary crafts some fake samples that the classifier cannot recognize correctly.  Poisoning attacks allow an attacker to insert or modify a small portion of fake samples in the training dataset. They aim to cause failures of the trained classifier such as poor accuracy or wrong prediction on some given test samples. Regarding the robustness of meta learning algorithms, recent works~\cite{yin2018adversarial} and~\cite{goldblum2019robust} show that the adapted model produced by meta learner is not robust toward adversarial examples. {\it Note that the goal of meta attack in this work is unique to meta learning that is different from that of ~\cite{yin2018adversarial,goldblum2019robust}.} It is more dangerous in the sense that we aim to destroy the overall performance of the adapted model without seeing and modifying the test samples.


 

\section{Meta Learning Basics and Notations}\label{meta}

In this section, we introduce key concepts, definitions and notations for meta learning that we will use in this paper. 
Generally, a meta-learning algorithm aims to learn a common strategy, i.e., a meta learner, to solve a variety of similar tasks $\mathcal{T}\sim p(\mathcal{T})$. In each task $\mathcal{T}$ (e.g. a few-shot classification task), a well trained meta learner $f_\theta$ with parameter $\theta$ utilizes the training data $D^{\mathcal{T}}_{\text{train}}$ to produce an adapted model $F(\cdot; \phi)$ with the parameter $\phi$. The adapted model is expected to have good prediction performance on the unseen test data $D_{\text{test}}^{\mathcal{T}}$. We denote such adaptation process as $\phi = f_\theta(D_{\text{train}}^{\mathcal{T}})$. 



To learn a good learning strategy, the meta learner $f_{\theta}$ needs to be trained on many similar tasks $\mathcal{T}\sim p(\mathcal{T})$ where $p(\mathcal{T})$ denotes the distribution of tasks. 
Specifically, there will be a training set $D_{\text{train}}^{\mathcal{T}}$ and a test set $D_{\text{test}}^{\mathcal{T}}$ associated with each task $\mathcal{T}$.
The goal of meta training is to learn a meta learner $f_{\theta}$ which is able to produce an adapted model $F(\cdot; \phi)$ incurring low test error on $D_{\text{test}}^{\mathcal{T}}$ across each task $\mathcal{T}$.
Formally, the objective function of meta training can be described as:
\begin{equation}
\label{equ:meta_traing}
    \setlength{\jot}{16pt}
    \begin{split}
            &\min_{\theta} \mathop{\mathbb{E}}_{\mathcal{T} \sim p(\mathcal{T})} \left(~~~\mathop{\mathbb{E}}_{x,y\sim D_{\text{test}}^{\mathcal{T}}} \mathcal{L} (F(x ; \phi), y) \right) \\
            &\text{   s.t.            }   \phi = f_{\theta}(D_{\text{train}}^{\mathcal{T}})
    \end{split}
\end{equation}
where $\mathcal{L}(y',y)$ denotes the loss with the prediction $y'$ and ground truth label $y$.  Note that in~\eqref{equ:meta_traing}, the parameters $\theta$ of the meta learner are the decision variable to be optimized. In practice, it is usually trained on a finite set of training tasks $\{\mathcal{T}_{\text{train}}^{i}, i = 1,2,...n_1\}$ while the performance of the trained meta learner will be evaluated on a finite set of test tasks $\{\mathcal{T}_{\text{test}}^{i}, i = 1,2,...n_2\}$. All notations and definitions are summarized in Table~\ref{Notations}.



\begin{table}[t]
\renewcommand{\arraystretch}{1.6}
\caption{Notations and Descriptions.}
\centering
\begin{tabular}{|c|c|}
\hline
Notation & Description \\
\hline\hline
$\mathcal{T}^i_{\text{train}}$ & Meta-train Task $i$ \\
\hline
$\mathcal{T}^i_{\text{test}}$ & Meta-test Task $i$ \\
\hline
$f_\theta(\cdot)$ & Meta learner with parameters $\theta$ \\
\hline 
$D^{\mathcal{T}}_{\text{train}}$ & Training data of task $\mathcal{T}$ \\
\hline
$D^{\mathcal{T}}_{\text{test}}$ & Test data of task $\mathcal{T}$ \\
\hline
$D^{\mathcal{T}}_{\text{adv}}$ & Adversarial data of  task $\mathcal{T}$ \\
\hline
$F(\cdot;\phi)$ & Adapted model with parameters $\phi$ \\
\hline
\end{tabular}
\label{Notations}
\end{table}

\section{Threat Model}\label{threat}

In this section, we describe details about the key components of our proposed threat model for meta learning including the victim model, adversary's goal and unnoticeable perturbation.

\subsection{Victim Meta Learner}\label{maml_model}


Generally, different meta learning algorithms have unique meta learner structures $f_{\theta}$. In this subsection, we briefly introduce three representative victim meta learners we attempt to attack in this work.


\subsubsection{Optimization-Based Meta Learner}
An optimization-based meta learner such as MAML~\cite{finn2017model} typically simulates the optimization process where the parameters of the adapted model $F(~\cdot~;\phi)$ are updated to achieve minimal loss on a task $\mathcal{T}$ with associated training data $D_{\text{train}}^{\mathcal{T}}$. 
For instance, MAML produces an adapted model $F(~\cdot~;\phi)$ by running $m$ gradient descent steps:
\begin{equation}
    \setlength{\jot}{15pt}
    \begin{cases}
         \phi_0 &= \theta - \alpha~\nabla_{\theta} \mathcal{L}(\theta)\\
        &...\\
        \phi_j &= \phi_{j-1} - \alpha~\nabla_{\phi_{j-1}} \mathcal{L}(\phi_{j-1})\\
         \phi &= \phi_m \text{~~~~~where~~~} m\geq 0
    \end{cases}
    \label{equ:maml_iter}
\end{equation}
where $\alpha$ is the step size and the parameters $\theta$ of the meta learner serve as the initialization of model $F(~\cdot~;\phi)$.
$\mathcal{L}(\phi_j)$ denotes the total training loss of model $F(\cdot;\phi_j)$ on $D_{\text{train}}^{\mathcal{T}}$: $$\mathcal{L}(\phi_j) = \sum_{(x,y)\in D_{\text{train}}^{\mathcal{T}}} \mathcal{L}(F(x;\phi_j), y).$$
By minimizing the training loss, the meta learner hopes to render a good model $F(\cdot, \phi)$ with small error on the test samples $D_{\text{test}}^{\mathcal{T}}$. 
The performance of the MAML model on a task $\mathcal{T}$ relies on the model initialization $\theta$ and the gradients. Later on, we will discuss how an attacker can manipulate the training samples to mislead MAML to update with ``malicious'' gradients.

\subsubsection{Model-Based Meta Learner}
Model-based meta learners such as MANN~\cite{santoro2016meta} and SNAIL~\cite{mishra2017simple} work as DNN models, which take inputs from $D_{\text{train}}^{\mathcal{T}}$ and directly output the adapted model $F(\cdot;\phi)$. For example, SNAIL~\cite{mishra2017simple} constructs an attention based sequential model which takes inputs from a series of samples $(x_1, y_1) ... (x_t,y_t)$ and outputs the label prediction for the last sample $(x_t, y_t)$. 
Specifically, the first $(t-1)$ input samples are from the training set $D_{\text{train}}^{\mathcal{T}}$, and it outputs the prediction for the $t$-th input, which is an upcoming test sample in $D_{\text{test}}^{\mathcal{T}}$.
For this type of meta learner, the training data $D_{\text{train}}^{\mathcal{T}}$ acts as parameters of the adapted model. Compared to other meta learning architectures, model-based meta learner is more similar to traditional DNN models. In our experiments, we choose the SNAIL~\cite{mishra2017simple} model as an example of the model-based meta learners to present its robustness behavior. 


\subsubsection{Metric-Based Meta Learner}

The metric-based meta learners are usually composed of two parts, a DNN model $g(\cdot,\theta)$ for feature extraction which projects all the training and test samples to a feature space, and a base classifier which divides the feature space to different classes. For example, in Prototypical Networks~\cite{snell2017prototypical}, when the meta leaner faces a task $\mathcal{T}$ with training set $D_{\text{train}}^{\mathcal{T}}$, in the embedding space, a nearest neighbor classifier is build on the embedded feature vectors \{$g(x;\theta) | x\in D_{\text{train}}^{\mathcal{T}}\}$. This classifier will make predictions for test samples by feeding them into the feature space and the base classifier. Take the Prototypical Net as an example, the robustness of this meta learner will be decided by both the DNN feature extractor and the base classifier. 


\subsection{Adversary's Goal}
In this work, the main goal of meta learning attacks (meta attack)
is to mislead the meta learner to produce ``malicious" models, which is different from traditional adversarial attacks focusing on test examples. Therefore, we need to formally redefine adversary's goals for meta attack. In this work, we study the robustness of meta learning for few-shot classification under the white-box setting where the adversary has the full knowledge of a trained meta learner, including the model parameters and adaptation process. We allow an adversary to manipulate a subset 
of the training set $D_{\text{train}}^{\mathcal{T}}$.
Under this constraint, the adversary can construct an adversarial training set $D_{\text{adv}}^{\mathcal{T}}$ which attempts to mislead the meta learner to produce a malicious adapted model. In particular, we consider two different types of adversarial goals to fool the meta learner including \textit{untargeted attack} and \textit{targeted attack}.
\subsubsection{\textbf{Non-targeted Attack} (Denial-of-Service Attack)}\label{untargeted}
In the setting of non-targeted attack, the adversary aims to let the adapted classifier $F(\cdot;\phi')$ 
have bad overall performance (low accuracy) across the test samples from the data distribution of the task $\mathcal{T}$. 
This adversarial objective can be formulated as to find an adversarial dataset $D_{\text{adv}}^{\mathcal{T}}$ that maximizes the test loss of the adapted model:
\begin{equation}
\label{equ:untargeted}
    \setlength{\jot}{14pt}
    \begin{split}
        \underset{D_{\text{adv}}^{\mathcal{T}}}{\text{maximize}}~~~&\mathop{\mathbb{E}}_{x,y\sim D_{\text{test}}^{\mathcal{T}}} \mathcal{L} (F(x ; \phi'), y)\\
        \text{s.t.~~~~        }   &\phi' = f_{\theta}(D_{\text{adv}}^{\mathcal{T}})         
    \end{split}
\end{equation}

Under the risk of untargeted attacks, the meta learner might build wrong or almost useless knowledge. 
This attack also called Denial-of-Service attack~\cite{wood2002denial}.

\subsubsection{\textbf{Targeted Attack}}\label{targeted}
Different from the untargeted attack which aims to degrade the overall performance for the task $\mathcal{T}$, the adversary under targeted attack takes certain subset of test samples $D_{\text{target}}^{\mathcal{T}} \subseteq D_{\text{test}}^{\mathcal{T}}$ as the targeted samples and aims to 
undermine the adapted model's performance on these targeted samples. 
Specifically, we consider all samples from one specific class as targeted samples in a task $\mathcal{T}$. In this way, we are able to evaluate whether the meta learner can be misled to produce an adapted model $F(\cdot;\phi)$ which has a general ``wrong'' concept for a given class. 
For example, a meta learner can build the knowledge about ``what is a cat'' after seeing a few teaching samples of cats. However, our adversary aims to let the adapted model misclassify any unseen cat images by attacking the meta learner. In this way, the adapted model's knowledge about the cat is destroyed. Formally, for a target class $t$, we define our targeted attack's objective as:

\begin{equation}
\label{equ:targeted}
    \setlength{\jot}{14pt}
    \begin{split}
        \underset{D_{\text{adv}}^{\mathcal{T}}}{\text{maximize}}~~~&\mathop{\mathbb{E}}_{x,y\sim D_{\text{test}, y = t}^{\mathcal{T}}} \left[\mathcal{L} (F(x ; \phi'), y)\right]\\
        \text{s.t.        }   &\phi' = f_{\theta}(D_{\text{adv}}^{\mathcal{T}})    
    \end{split}
\end{equation}

In fact, we empirically show that an attacker can break the adapted model's knowledge on one class even by only modifying teaching samples in another class. For example, an attacker perturbs a few samples of dog images, but the adapted classifier can fail to recognize cat images. This phenomenon provides strong evidence on the unreliability of meta learning models. More details can be found in the experiment section. 

\subsubsection{\textbf{Surrogate Test Loss}}
In the objectives of untargeted attacks~\eqref{equ:untargeted} and targeted attacks~\eqref{equ:targeted}, the attacker needs to have knowledge about test samples $D_{\text{test}}^{\mathcal{T}}$ which is not realistic in  real world scenarios.
Therefore, we propose to use the empirical training loss on training samples $D_{\text{train}}^{\mathcal{T}}$ to approximate the test loss. When doing meta adversarial attack, we hope our adapted model from perturbed train set can generalize the ``malicious'' knowledge to unseen test samples.  
Formally, we unify the untargeted and targeted objectives as follows:
\begin{equation}
\setlength{\jot}{12pt}
    \begin{split}
         &\underset{D_{\text{adv}}^{\mathcal{T}}}{\text{maximize}}~~~\sum_{x,y\in D_{\text{train}}^{\mathcal{T}}} \left[\mathcal{L} (F(x ; \phi'), y)\right]\\
        &\text{s.t.        }   \phi' = f_{\theta}(D_{\text{adv}}^{\mathcal{T}}) 
    \end{split}
\end{equation}
where $D_{\text{train}}^{\mathcal{T}}$ is actually $D_{\text{train}, y=t}^{\mathcal{T}}$ in the targeted case. 
The generalization of attack from $D_{\text{train}}^{\mathcal{T}}$ to $D_{\text{test}}^{\mathcal{T}}$ further reveals the unreliability of meta learning as showed in the experiments.

\subsection{Unnoticeable Perturbation}\label{unnotice}
Typically, in an adversarial attack scenario, the attacker is allowed to modify the input data in a sneaky and unnoticeable way. 
With the goal of meta attack, the unnoticeable perturbation is also a valid concern but how to define unnoticeable perturbations in such setting has not been established yet.
In this work, we provide two principles, i.e., perturbed samples budget and perceptual similarity, to ensure that the perturbed dataset $D_{\text{adv}}^{\mathcal{T}}$ is similar to $D_{\text{train}}^{\mathcal{T}}$. 

\subsubsection{\textbf{Perturbed Samples Budget}} The adversary is required to perturb as few samples as possible to achieve the adversarial goal because fewer fake samples an adversary injects to the system, the less likely this attack can be detected. 
We denote $x^{\text{adv}}\in D_{\text{adv}}^{\mathcal{T}}$ as a perturbed sample in the perturbed set $D_{\text{adv}}^{\mathcal{T}}$ and $x=\text{clean}(x^{\text{adv}})$ as the corresponding clean sample. Formally, we require the perturbed samples budget is limited by $k$:
\begin{equation}
    \sum_{x^{\text{adv}}\in D_{\text{adv}}^{\mathcal{T}}} ~~~\mathbb{1}(x^{\text{adv}} \neq \text{clean}(x^{\text{adv}}) )\leq k
\end{equation}

\subsubsection{\textbf{Perceptual Similarity}} In each individual perturbed sample, we require that the perturbed image is perceptually similar to the clean image. In other words, our added perturbation is indistinguishable for human. We manage to achieve this criterion by limiting the perturbation as follow:
\begin{equation}
        ||x^{\text{adv}} - \text{clean}(x^{\text{adv}})|| \leq \epsilon, ~~\forall x^{\text{adv}} \in D_{\text{adv}}^{\mathcal{T}}. 
\end{equation}


\footnotetext{In image domain, we usually use $l_p$ norm difference to define the perceptual dissimilarity between two images. In the future parts of this work, we implicitly denote $||\cdot||$ as $l_\infty$ norm.}

\section{Meta Attack}
\label{sec:alg}
Based on the adversary's goal and capacity as described in the last section, the meta attack problem can be defined as the follows:

\newtheorem{prob}{\textbf{Problem}}
\begin{prob}
    Given a well-trained 
    meta learner $f_\theta$, a new unseen learning task $\mathcal{T}$, the corresponding training samples $D_{\text{train}}^{\mathcal{T}}$ and perturbation budget $(k,\epsilon)$,
    we aim to find an adversarial training set $D_{\text{adv}}^{\mathcal{T}}$ by solving the following optimization problem:
    \setlength{\jot}{12pt}
    \begin{align}\label{eq:prob1}
            \setlength{\jot}{12pt}
            \begin{gathered}
            \underset{D_{\text{adv}}^{\mathcal{T}}}{\text{maximize}}~~~\sum_{x,y\in D_{\text{train}}^{\mathcal{T}}} \left[\mathcal{L} (F(x ; \phi'), y)\right]\\
            \text{s.t.~~ } 
            \begin{cases}
                & \phi' = f_{\theta}(D_{\text{adv}}^{\mathcal{T}}) \\ 
                \noalign{\vskip8pt}
                & \sum_{x^{\text{adv}} \in D_{\text{adv}}^{\mathcal{T}}} ~~~ \mathbb{1}(x^{\text{adv}} \neq \text{clean}(x^{\text{adv}}))\leq k, \\
                \noalign{\vskip8pt}
                & || x^{\text{adv}} - \text{clean}(x^{\text{adv}})|| \leq \epsilon, ~~~ \forall x^{\text{adv}} \in D_{\text{adv}}^{\mathcal{T}}.   
            \end{cases}
            \end{gathered}
    \end{align}
\end{prob}
In other words, in Problem $1$, we aim to perturb at most $k$ samples within training data set $D_{\text{train}}^{\mathcal{T}}$ with $\ell_{\infty}$ norm perturbation constraint $\epsilon$ such that the resulting adversarial data set $D_{\text{adv}}^{\mathcal{T}}$ attacks the meta learner $f_{\theta}$ which unexpectedly produces an adapted model $F(\cdot;\phi')$ incurring maximal loss value on the targeted samples $D_{\text{train}}^{\mathcal{T}}$. Selecting the samples to be perturbed in $D_{\text{train}}^{\mathcal{T}}$ is a combinatorial optimization problem and we provide a greedy algorithm for such selection process as showed in Alg.~\ref{alg:search}. Before introducing that, we first describe the meta attacking algorithm given a selected set 
$D_{\text{select}}^{\mathcal{T}} \subset D_{\text{train}}^{\mathcal{T}}$ 
as showed in Alg.~\ref{alg:adver}.


\begin{algorithm}[h]
\begin{algorithmic}[1]
\setstretch{1.3}
 \STATE \textbf{Input:} A new task $\mathcal{T}$ and the associated training set $D_{\text{train}}^{\mathcal{T}}$, a given selected set $D_{\text{select}}^{\mathcal{T}}$, 
 $\ell_\infty$ norm constrain $\epsilon$ and attacking step $K$  \\
 \STATE \textbf{Output:} adversarial sample set $D_{\text{adv}}^{\mathcal{T}}$
 \STATE Initialize $x_i^0 = x_i,~\forall x_i \in D_{\text{select}}^{\mathcal{T}}$ \\
\FOR{$k=0,\dots, K-1$}
\STATE $D_{\text{adv}}^{\mathcal{T}} = \{ D_{\text{train}}^{\mathcal{T}} \cap D_{\text{select}}^{\mathcal{T}} \} \cup D_{\text{select}}^{\mathcal{T}}$ \\ 
\STATE $\phi^k = f_{\theta}(D_{\text{adv}}^{\mathcal{T}})$ \label{alg:newmodel} \\
\STATE $\mathcal{L}_{\text{total}}(\phi^k) = \sum_{(x,y)\in D_{\text{train}}^{\mathcal{T}}} \mathcal{L}(F(x,\phi^k), y)$
\FOR{$x_i^k \in D_{\text{select}}^{\mathcal{T}}$}
    \STATE $g_i^{k} = \nabla_{x_i^{k}} \mathcal{L}_{\text{total}}(\phi^k)$ \\
    \STATE $x_i^{k+1} = \text{Clip}_{x_i, \epsilon} \left ( x_i^{k} + \alpha~\text{sign} (g_i^{k})\right )$ \label{alg:pgdstep}  \\
\ENDFOR
\ENDFOR
 \STATE \textbf{Return}: $D_{\text{adv}}^{\mathcal{T}}$
\caption{Generate adversarial set for a given selected set}
\label{alg:adver}
\end{algorithmic}
\end{algorithm}

\subsection{Generating Adversarial Samples for A Given Selected Set}

In each iterative step, the adversarial sample set $D_{\text{adv}}^{\mathcal{T}}$ is first constructed by replacing the corresponding clean samples in $D_{\text{train}}^{\mathcal{T}}$ with $D_{\text{select}}^{\mathcal{T}}$ (line 5).
Based on $D_{\text{adv}}^{\mathcal{T}}$, the meta learner produces an adapted model $F(\cdot;\phi^k)$ with parameters $\phi^k = f_{\theta}(D_{\text{adv}}^{\mathcal{T}})$ (line~\ref{alg:newmodel}). A total loss value $\mathcal{L}_{\text{total}}(\phi^k)$ of model $F(\cdot;\phi^k)$ is computed based on the clean training set $D_{\text{train}}^{\mathcal{T}}$ (line 7).
To maximize this loss, we perturb each sample $x_i^k$ in the selected set through a projected gradient ascent step (line 9 and 10). The \textit{Clip} function denotes the projection function which projects the perturbed sample $x_i^k$ to the $\epsilon$-neighborhood of clean sample $x_i$, i.e., $B_{\epsilon}(x_i):\left\{x':||x'-x_i||\leq \epsilon\right\}$. 
The gradient is calculated with respect to the sample $x_i^k$ through the chain rule:
\begin{align}
    \nabla_{x^k_i} \mathcal{L}_{\text{total}}(\phi^k) = \frac{\partial \mathcal{L}_{\text{total}}(\phi^k)}{\partial \phi^k} \frac{\partial \phi^k}{\partial x^k_i} 
    = \frac{\partial \mathcal{L}_{\text{total}}(\phi^k)}{\partial \phi^k} \frac{\partial f_\theta(D_{\text{adv}}^{\mathcal{T}})}{\partial x^k_i}.
\end{align}
Note that the computation of $\frac{\partial \mathcal{L}_{\text{total}}(\phi^k)}{\partial \phi^k}$ is relatively easy but the computation of the Jacobian matrix $\frac{\partial f_\theta(D_{\text{adv}}^{\mathcal{T}})}{\partial x^k_i}$ is much more challenging because
the adapted model $\phi^k$ has complicated and distinct dependency on sample $x_i^k \in D_{\text{adv}}^{\mathcal{T}}$ according to the structure of meta learner $f_{\theta}$.

\begin{figure}[h]
\centering
  \includegraphics[width=0.9\linewidth]{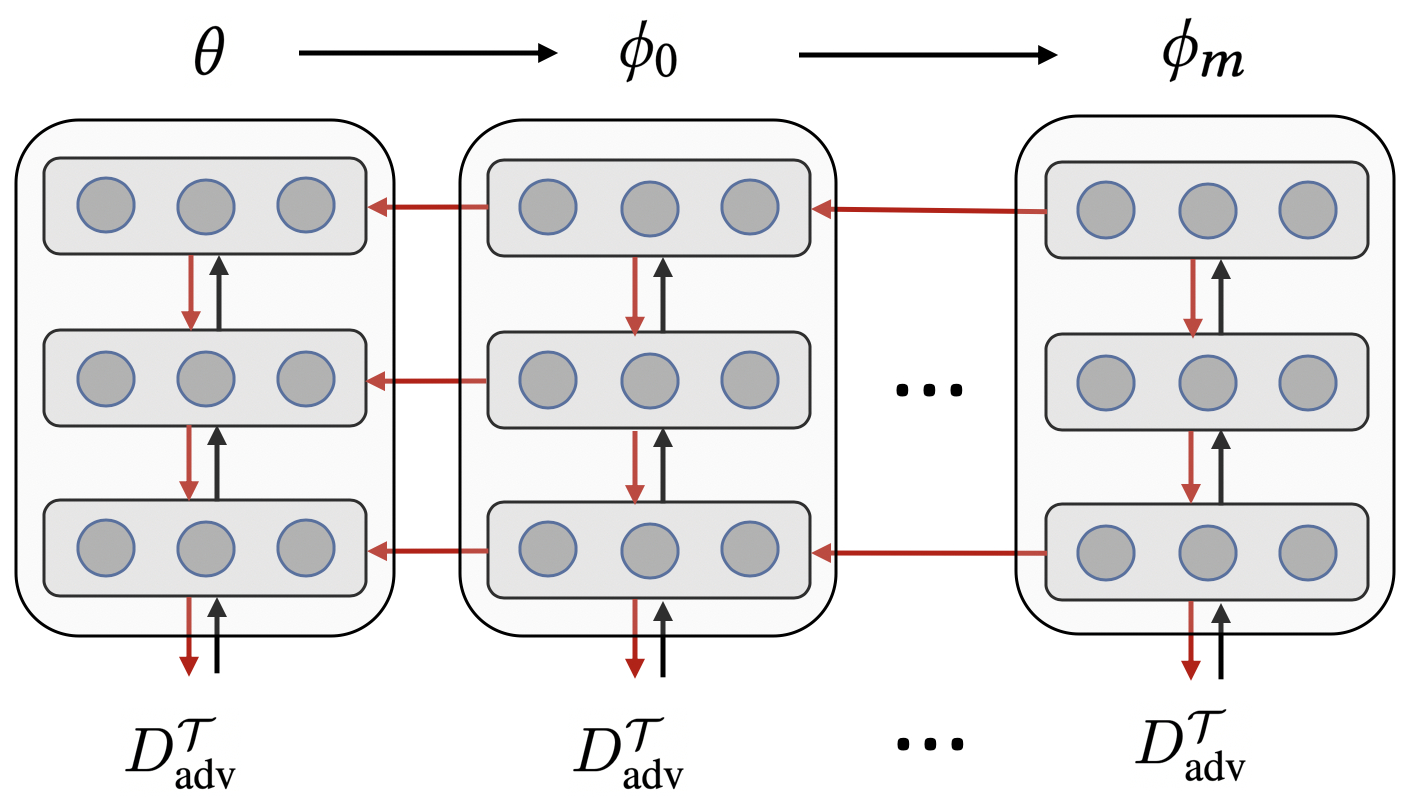}
  \caption{
  Computational graph for Jacobian matrix computation $\frac{\partial f_\theta(D_{\text{adv}}^{\mathcal{T}})}{\partial x^k_i}$
  The red arrows indicate the backward propagation through the meta learning adaptation steps.}
\label{fig:gradient}
\end{figure}

To provide more insights, we take Model-Agnostic Meta Learning (MAML) model~\cite{finn2017model} as an example. Taking the input data set $D_{\text{adv}}^{\mathcal{T}}$, MAML will produce a new model $\phi = \phi_m$ after $m$ iterative gradient updates as described in~\eqref{equ:maml_iter}. Therefore, 
the adapted model $f_{\theta}(D_{\text{adv}}^{\mathcal{T}})$
will have dependencies on the input data $x_i \in D_{\text{adv}}^{\mathcal{T}}$ through each intermediate model $\phi_j$ where $j=0,1,\dots,m$ as showed in the computation graph in Figure~\ref{fig:gradient}. 
As a result, the Jacobian matrix is calculated through the backward propagation from $\phi_m$ to each input $x_i$ in each step $j$ and the final result will be the summation of matrices calculated from each backward path. Computationally, this requires backward passes through higher order derivatives, which is supported by standard deep learning libraries such as PyTorch \textit{autograd}~\cite{NEURIPS2019_9015}.




This higher-order gradient attacking algorithm will help adversarial examples to mislead the MAML model to produce ``malicious'' parameter updates for multiple gradient steps. We note that this higher order gradient calculation is necessary and as the MAML implements more updating steps, one-order gradient attacks cannot result in significant influence on the adapted model. In our experimental results, we will show that the our proposed algorithm can effectively decrease the MAML's robustness with multiple updating steps.

Different from MAML, for SNAIL~\cite{mishra2017simple} and Prototypical Networks~\cite{snell2017prototypical}, where the meta learner is one DNN model or the combination of a DNN + Nearest-Neighbor classifier, one-order gradient attack is sufficient to find the most influential adversarial set.


\subsection{Search for the Optimal Adversarial Set}

In corresponding to the definition of unnoticeable perturbation in Section~\ref{unnotice}, we will need to constrain the perturbation budget to only select at most $k$ samples to perturb in $D_{\text{train}}^{\mathcal{T}}$. Note that how to choose the most influential set for adversarial attack also can affect the attacking performance. 
However, the searching process will be costly if the data set $D_{\text{train}}^{\mathcal{T}}$ is large because it requires to search over $C(|A|, k)$ combinations. Therefore, we provide a greedy algorithm to obtain an approximate solution to keep adding the most dangerous adversarial sample into the attacking bag. It is briefly shown in Algorithm~\ref{alg:search}. In each iteration $i$, we choose one sample $x_{\text{opt}}$ from $D_{\text{train}}^{\mathcal{T}} \backslash S_{i-1} $ which incurs the largest adversarial loss $\mathcal{L}_{\text{total}}$ when added into the set $S'$. 
In this way, we iteratively enlarge our candidate set by constructing the most adversarial $1$-set, $2$-set until $k$-set perturbation. 

\SetKwInput{KwInput}{Input}                
\SetKwInput{KwOutput}{Output}              
\begin{algorithm}[h]
\setstretch{1.3}  
  \KwInput{Clean train set $D_{\text{train}}^{\mathcal{T}}$, 
  perturbed sample budget $k$, and $\ell_\infty$ norm constrain $\epsilon$}
  \KwOutput{Perturbation Set $D_{\text{select}}^{\mathcal{T}}$}
  Initialize $i\leftarrow 0$, 
  $S_0 = \emptyset$

   \While{$i< k$}
   {
    i = i+1 \\
   	\For{each $x \in D_{\text{train}}^{\mathcal{T}} \backslash S_{i-1} $}{
    $S' = \{x\} \cup S_{i-1}$ \\
    $D_{\text{input}}^{\mathcal{T}} = \{D_{\text{train}}^{\mathcal{T}} \cap S'\} \cup S'$ \\
    Generate adversarial perturbation on set $D_{\text{input}}^{\mathcal{T}}$ with Alg.~\ref{alg:adver} and store the adversarial loss $\mathcal{L}_{\text{total}}$
    }
    Choose the sample $x_{\text{opt}}$ which incurs largest $\mathcal{L}_{\text{total}}$ \\ 
    $S_i = S_{i-1} \cup \{x_{\text{opt}}\}$
   }
  \textbf{Return}: $S_{k}$
\caption{Search the perturbation set}
\label{alg:search}
\end{algorithm}

\setstretch{1}

\section{Experiment}\label{exp}

In this section, we evaluate the proposed meta attack algorithm MetaAttacker against three popular meta learning algorithms, including MAML~\cite{finn2017model}, SNAIL~\cite{mishra2017simple} and Prototypical Networks~\cite{snell2017prototypical}, under different settings as we introduced in Section~\ref{threat}. Through the experiments, we aim to answer the following questions: 
(1) Can we successfully attack the meta learner by inserting unnoticeable perturbations under different perturbation budgets? 
(2) How do the hyperparameters of meta learner influence its robustness and reliability against meta learning attack? and (3) Will different meta learners present different robustness behaviors to adversarial attacks? Next, we first discuss the full results on MAML~\cite{finn2017model} model to gain an overview understanding on its robustness under different settings. Then, we implement our attacks on SNAIL~\cite{mishra2017simple} and Prototypical Networks ~\cite{snell2017prototypical} to study the difference between different meta learning structures. 

\subsection{Experimental Setup}

\textbf{Datasets.} To evaluate the performance and robustness of existing meta-learning algorithms, we apply the proposed attacking framework to few shot learning problems on two the most common used benchmark datasets including Omniglot~\cite{lake2015human} and MiniImagenet~\cite{ravi2016optimization} datasets. The Omniglot dataset consists of 1,623 human-writing characters where each character has 20 different images. The MiniImagenet dataset consists of 100 classes with 600 samples of 84×84 color images per class. For both datasets, we report meta learning performance and robustness performance under 5-way 5-shot classification problems. In each 5-way 5-shot learning task, we have 25 training samples in total (5 samples per class) and use 15 test samples per class to calculate the test accuracy. In the evaluation phase, we report the average test accuracy across 100 meta-test tasks $\mathcal{T}_i = \{D_{\text{train}}^{\mathcal{T}_i}, D_{\text{test}}^{\mathcal{T}_i}\}~~, i = 1,2,...,100$.

\textbf{Unnoticeable Perturbation} For the Omniglot dataset, it consists of handwritten character images by pixel resolution $28\times 28$ in the range $[0,1]$, which is similar to MNIST~\cite{lecun1998gradient}. Thus, we define the perturbation in this dataset to be unnoticeable by constraining $l_\infty$ norm not larger than $0.3$: $||x^{\text{adv}}-x||\leq 0.3$. For Mini-Imagenet dataset whose image size is $84\times84$, we constrain the unnoticeable perturbation by limiting $||x^{\text{adv}}-x||\leq 8/255$. Generally, for a 5-way 5-shot classification problem, we will limit the perturbation budget where the attacker cannot attack more than 1,2,3 or 5 images out of these 25 images.

\subsection{Experimental Results for MAML}

\subsubsection{\textbf{Clean Performance.}} 

For an illustration of our attack on MAML, we choose the backbone model structure as a DNN model with 4-layer Convolutional Neural Network as used by~\cite{finn2017model} and \cite{vinyals2016matching}.  As suggested by \cite{finn2017model}, fine-tuning step $m$ on a local training set $D_{\text{train}}^{\mathcal{T}}$ is one important hyperparameter that influences the meta learner's performance.
Therefore, in Table~\ref{table:2}, we report the model's clean performance under different settings of fine-tuning steps ($m = 1,5,10$). From the clean performance in Table~\ref{table:1}, we can see that more fine-tuning steps will help to improve the clean performances of MAML in both Omniglot and Mini-ImageNet.

\begin{table}[h]
\centering
\caption{Average test accuracy (in $\%$) for MAML on Omniglot and Mini-Imagenet. We list the performance for 1,5,10 fine-tuning steps of MAML adaptation during the test phase.} 
\label{table:1}
\begin{tabular}{c|ccc}
\toprule
F.T Step  &1 Step & 5 Step  & 10 Step\\
\midrule
Omniglot  & 99.0 & 99.9 & 99.9\\
Mini-Imagenet & 58.8 & 62.6 & 63.1\\
\bottomrule
\end{tabular}
\end{table}

\subsubsection{\textbf{Non-targeted Attack Performance}}
We study the performance of our non-targeted attacking algorithm which aims to influence overall accuracy of the adapted classification model as described in Section~\ref{untargeted}. We evaluate the adapted model's average accuracy across 100 test tasks \{$\mathcal{T}_i, i = 1,2,...,100$\}. For each single task $\mathcal{T}$, the adversary has the access to manipulate all training images, which means that the accessible set $A = D_{\text{train}}^{\mathcal{T}}$. However, it is only allowed to modify no more than $k = 1,2,3,10$ samples in $D_{\text{train}}^{\mathcal{T}}$, 

In order to validate the effect of our adversarial attack, we compare our meta attack algorithm versus adding random perturbations in the constrained unnoticeable space $\{x^{\text{rand}}:||x^{\text{rand}}-x||\leq \epsilon\}$, on random chosen samples of the training set. In our experiments, we set the attacking step size $\alpha = 0.01$ for Omniglot, $\alpha = 0.2/255$ for Mini-Imagenet, and attacking steps length $l = 100$. We report the average test accuracy across 100 meta-test tasks. The detailed experimental results on Mini-Imagenet and Omniglot are shown in Table~\ref{table:2}. 

For the Mini-Imagenet dataset, the MAML meta-learner has a clean performance with average accuracy around $60\%$, which means there are some tests that the adapted classifier has low accuracy (smaller than $40\%$). It is meaningless to attack the cases where meta learner fails. Thus, in our experiments, we select the successfully adapted test tasks in which the mini test accuracy is over $55\%$. In addition to the attacking results under different perturbation budgets, we also present two baseline performance. \textit{Non-attack} denotes the clean test performance of MAML across all the selected tasks. \textit{Random F.T.} means that for each task $\mathcal{T}$, we randomly initialize the model parameters $\phi_0$, and do fine tuning from this randomized $\phi_0$. Since MAML is essentially about to find a proper initialization for task $\mathcal{T}$, we use \textit{Random F.T.} to show the situation where the learning process has no guide from MAML.


\begin{table}[h]
\centering
\caption{Average testing accuracy of MAML adapted models after MetaAttacker attack vs. Random attack on Mini-Imagenet and Omniglot, when MAML takes 1, 5, 10 steps fine-tuning. The perturbation budget is under 1, 2, 5 and 10 samples respectively.}
\label{table:2}
Mini-Imagenet Dataset\\
\resizebox{0.5\textwidth}{!}{
\begin{tabular}{c|cc|cc|cc}
\toprule
F.T. Step &  \multicolumn{2}{c|}{1 Step} & \multicolumn{2}{c|}{5 Step} & \multicolumn{2}{c}{10 Step}\\
Attack Method  & MetaAttacker   & Rand.    & MetaAttacker   & Rand.  &MetaAttacker  & Rand. \\
\hline
1 Sample   &  48.2 &  63.0  & 55.6  & 64.7  & 56.6 & 66.3\\
2 Samples  &  42.6 & 62.9   & 44.6  & 64.4  &51.2  &  66.2\\
5 Samples   &  25.6  &  62.9   & 33.0  & 64.6  & 40.6 & 65.7\\
10 Samples   &  16.2  &  62.2   & 20.4  & 64.1  & 22.8 & 65.4\\
\hline\hline
Non-attack   &  \multicolumn{2}{c|}{63.3}  &  \multicolumn{2}{c|}{64.8} &  \multicolumn{2}{c}{66.5}\\
Random F.T.   &  \multicolumn{2}{c|}{23.4}  &  \multicolumn{2}{c|}{23.4} &  \multicolumn{2}{c}{27.5}\\
\bottomrule
\end{tabular}
\vspace{0.4cm}
}
\vskip 0.2in
Omniglot Dataset\\
\resizebox{0.5\textwidth}{!}{
\begin{tabular}{c|cc|cc|cc}
\toprule
F.T. Step &  \multicolumn{2}{c|}{1 Step} & \multicolumn{2}{c|}{5 Step} & \multicolumn{2}{c}{10 Step}\\

Attack Method   & MetaAttacker   & Rand.    & MetaAttacker   & Rand.  &MetaAttacker  & Rand. \\
\hline
1 Sample& 96.2& 99.0& 99.2& 99.8& 99.3 & 99.9 \\
2 Samples& 93.3& 98.8& 96.9& 99.3& 96.6 & 99.2 \\
5 Samples& 77.8& 98.4& 82.2& 99.1& 80.9 & 99.2 \\
10 Samples& 61.4& 98.1& 63.8& 98.9& 62.4 & 99.0 \\
\hline\hline
Non-attack   &  \multicolumn{2}{c|}{99.0}  &  \multicolumn{2}{c|}{99.8} &  \multicolumn{2}{c}{99.9}\\
Random F.T.   &  \multicolumn{2}{c|}{46.6}  &  \multicolumn{2}{c|}{52.6} &  \multicolumn{2}{c}{54.4}\\
\bottomrule
\end{tabular}
}
\end{table}

From Table~\ref{table:2}, we note that generating random noise on random samples can hardly influence MAML's performance. For our proposed MetaAttacker attack on Mini-Imagenet, the most-successful attacking case (modifying 10 samples when MAML takes 1 step fine-tuning) reduces the average accuracy from $63.3\%$ to $16.2\%$. The most-difficult attacking setting (modifying 1 sample under 10 steps fine-tuning) can also reduce overall accuracy from $65.2\%$ to $56.6\%$.  For each case, when there are about 10 out of 25 samples are modified, the MAML adapted models' performance is similar or below \textit{Random F.T.}, which suggests that the guide or experience from MAML under adversarial attacks becomes similar to randomized initialization. On the Omniglot dataset, MAML meta learner for 5-way 5-shot learning tasks can achieve very high performance on clean inputs $D_{\text{train}}^{\mathcal{T}}$ (almost $99\%$ average test accuracy across test tasks). From the results on Table \ref{table:2}, we can see that an adversary is required to perturb at least 2 samples to reduce the meta learner performance by $2\sim5\%$, and 5 samples to reduce the performance by $20\%$. 
We note that one intriguing property of the robustness of MAML is that with the increase of the fine tuning steps, the meta learner's robustness is also improved.

\subsubsection{\textbf{Targeted Attack Performance}}
In this subsection, instead of observing the meta learner's overall robustness performance, we focus on studying the robustness of meta learner via a local view which locates to one single class for each learning task. Specifically, we consider the setting where the adversary aims to influence the model's prediction on test samples from a specific target class: $y=\text{target}$ (recall our discussion in Section~\ref{targeted}). Under this adversarial goal, we further constrain that the adversary is only allowed to manipulate training samples in one given class $y=\text{attack}$. In our experiments, we consider the following two settings -- (1) \textbf{Direct Attack}: the adversary can manipulate samples from the target class, i.e., \textit{target} = \textit{attack}; and (2) \textbf{Influence Attack}: the adversary can only manipulate train samples from a different class, i.e., \textit{target} $\neq$ \textit{attack}. By studying meta learner's manner under the targeted attack, we can hopefully study the meta learner's safety manner in an inter-class level. For example, we try to answer the question: can we perturb the training samples in one class, but make the adapted model hardly recognize samples in another class? Is it easier to attack the adapted model performance when the adversary directly inserts perturbation into the target class? In our experiments, to evaluate the classification models' performance on a specific target class, for each mini task $\mathcal{T}$, we list the ``in class'' accuracy of the target class, or namely ``recall'' of the target class. The recall represents the percentage of the correctly classified  test samples among all test samples in the targeted class.

\begin{figure*}[h]
\centering
\subfloat[]{
\begin{minipage}[t]{0.33\textwidth}
\includegraphics[width=1.0\textwidth]{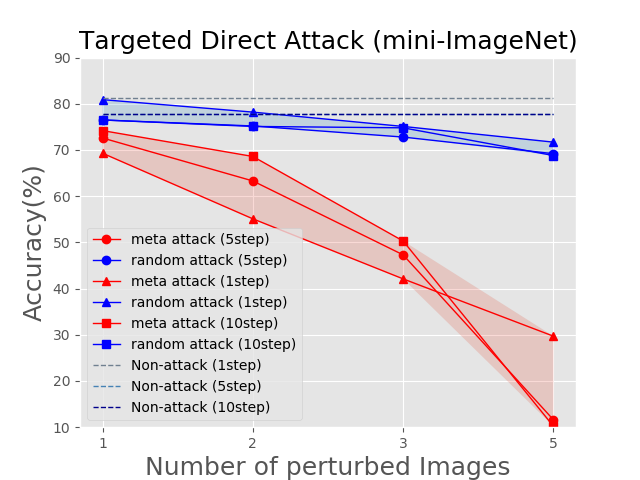}
\end{minipage}
}
\subfloat[]{
\begin{minipage}[t]{0.33\textwidth}
\includegraphics[width=1.0\textwidth]{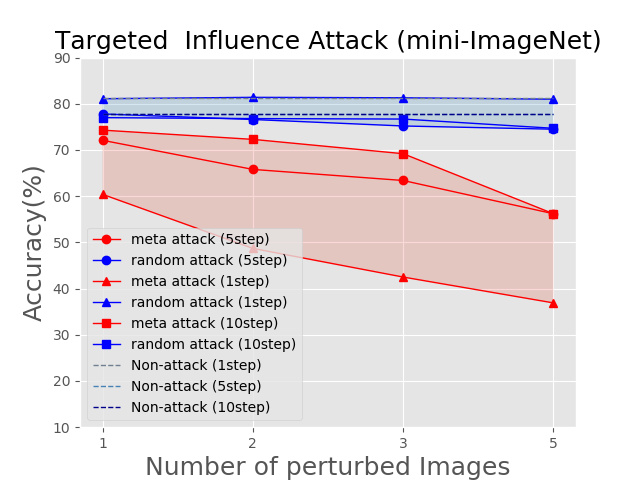}
\end{minipage}
}
\vspace{-0.1in}
\\
\subfloat[]{
\begin{minipage}[t]{0.33\textwidth}
\includegraphics[width=1.0\textwidth]{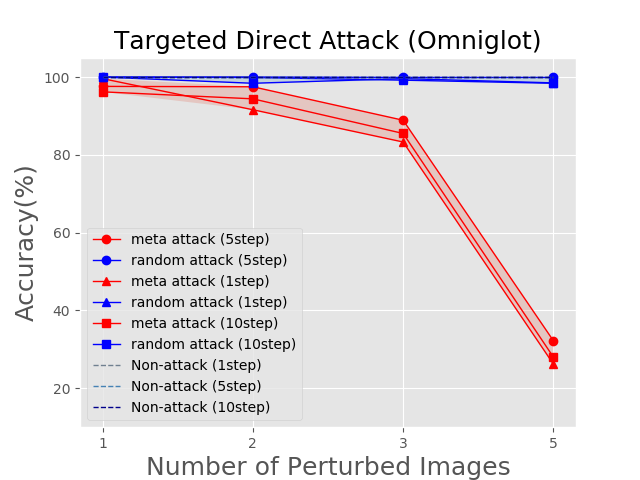}
\end{minipage}
}
\subfloat[]{
\begin{minipage}[t]{0.33\textwidth}
\includegraphics[width=1.0\textwidth]{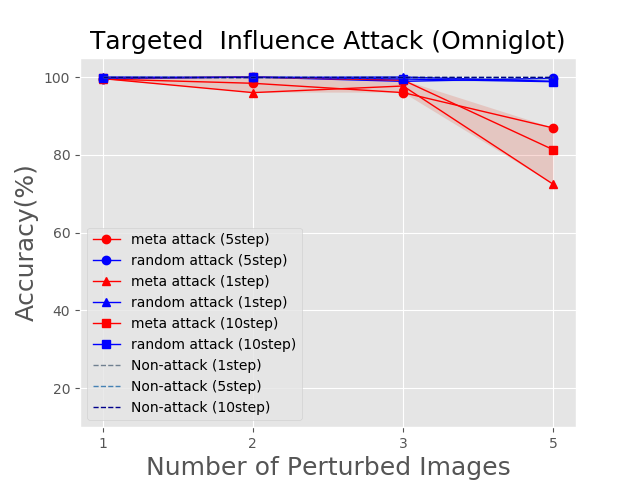}
\end{minipage}
}
\caption{Targeted attack performance on MAML meta learner, on Mini-Imagenet (top) and Omniglot (bottom)}
\label{fig:compare}
\end{figure*}








The results are shown in Figure~\ref{fig:compare}. From the figure, we can see that both direct attack and influence attack have achieved the adversarial goal to influence the adapted models' performance on target classes. Overall, the direct attacks are more powerful than influence attacks on most cases when the meta learner takes 5 steps and 10 steps fine tuning, especially when the allowed perturbation budgets are respectively large. When the adversary is allowed to manipulate all 5 training samples in the target class, he can reduce the in-class accuracy to around $10\%$, so the model can hardly recognize the test samples in this targeted class. Remark that even though we perturbed all 5 train samples in the target class, we still constrain the perceptual similarity for them. When the adversary is not allowed to directly perturb the train samples in the target attack (influence attack), he can still mislead the meta learner to have decreased accuracy for samples in the target class. This attacking effect is most obvious when the meta learner takes 1 step fine tuning. However, when meta learner takes more steps of fine tuning, the effect from direct attack is more powerful than influence attack.

\begin{figure*}[!h]
\centering
\subfloat[]{
\begin{minipage}[t]{0.32\textwidth}
\label{fig:more-untarget}
\includegraphics[width=1.0\textwidth]{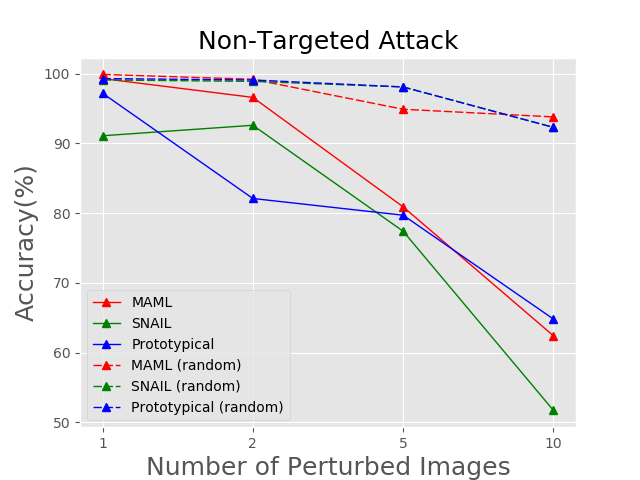}
\end{minipage}
}
\subfloat[]{
\begin{minipage}[t]{0.32\textwidth}
\label{fig:more-directtarget}
\includegraphics[width=1.0\textwidth]{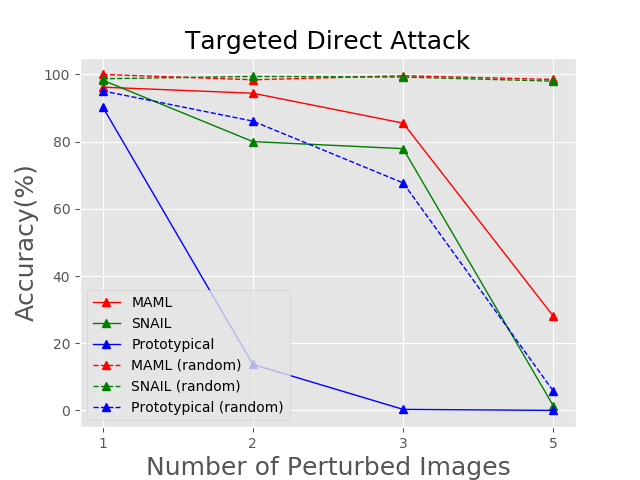}
\end{minipage}
}
\subfloat[]{
\begin{minipage}[t]{0.32\textwidth}
\label{fig:more-influencetarget}
\includegraphics[width=1.0\textwidth]{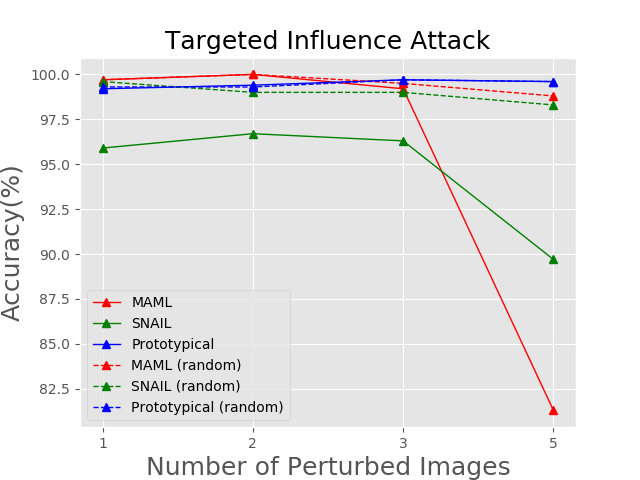}
\end{minipage}
}
\caption{Attack Performance on Different Meta Learning Models.}
\end{figure*}

\subsection{Attacking Other Meta Learning Models} \label{other}



In addition to MAML, we also consider two other types of meta learning models, including one model-based meta learner SNAIL~\cite{mishra2017simple} and one metric-based meta learner Prototypical Networks~\cite{snell2017prototypical}. The sequential model for SNAIL is inherited from the setting of the original work~\cite{mishra2017simple}, which contains two blocks of Temporal Convolutional layers and two causal attention layers.
The DNN model structures of Prototypical Networks we are using are also inherited from the original work in \cite{snell2017prototypical}, which are composed of 4 convolutional blocks with each block containing a 64-filter 3×3 convolution. For empirical study for the meta learner's robustness, we report their non-targeted and targeted attacked performance in Omniglot dataset since we have similar observations on Mini-Imagenet. As shown in Figure~\ref{fig:more-untarget},  both these two meta learner models are also vulnerable to non-targeted adversarial attacks, and with the increase of perturbation budgets, the average performance drops significantly. While, for targeted attacks as shown in Figures~\ref{fig:more-directtarget} and~\ref{fig:more-influencetarget}, the SNAIL models are easy to be attacked by both direct and influence attacks. However, for Prototypical Networks, the direct attacks can result in huge performance degradation for the target samples, but influence attacks can hardly have impacts on the target samples.

\subsection{Discussion}

Traditionally, in classification tasks, our ``perturbing train set to influence test performance'' is a data poisoning process. For the non-targeted attack (Denial-of-Service attack), there are existed data poisoning attacks~\cite{biggio2012poisoning, mei2015using}, which concentrate on traditional machine learning models, such as linear Support Vector Machine and linear regression models. For examples, \cite{biggio2012poisoning} generate clean poison on one training sample of MNIST dataset, which increase $5\% \sim 15\%$ error rate. In the deep learning scenario, the data poisoning becomes much more difficult because deep neural networks are usually trained on large datasets with many training epochs. Thus, only poisoning a few samples can hardly have large effect on the final trained model. We note that there are some existing works, such as~\cite{biggio2012poisoning, koh2017understanding} which aim to let the trained model have wrong prediction on a small set of test samples. For example, in~\cite{koh2017understanding}, the authors manage to flip the trained Inception~\cite{szegedy2015going} model's prediction on one targeted test sample from Imagenet with $57\%$ successful rate by perturbing one random training sample. 

For our proposed adversarial attack on meta learning, our attacking can largely change the adapted DNN model's overall performance on unseen test samples by only perturbing a few training images. We also show that attacking meta learner can alter the adapted DNN's prediction on all unseen test samples in one targeted class. The generalization of attacking effect also uncovers the vulnerability of meta learning algorithms from a new perspective. Therefore, we can conclude that the DNN model adapted from meta learning process is in more danger under the risk of data poisoning. It is because the meta learner only adapts on a few training samples, and has a few tuning steps. The given meta learner structure will also explicitly guide an attacker to insert the poisoned samples.

\section{Conclusion}
\label{sec:conclusion}

In this work, we first formally define the adversarial attacks and robustness issues for meta learning algorithms. We emphasize the existence of meta adversarial examples can be dangerous in real world scenarios. In detail, we give new definitions for adversarial goal and unnoticeable perturbations for attacking meta learning algorithms. Based on our definition for adversarial attacks for meta learning, we design efficient attacking approaches to fulfill our goal and validate our approach on different datasets for various meta learning models. Our empirical studies show that the attack method can result in significant performance drop for these meta learning models. This study opens doors to the security issues about meta learning. There are many directions needing further investigations. First, our current attacking strategy focused on the white-box setting and we would like to study strategies for other settings such as the black-box setting. Second, we aim to derive extensions of existing meta learning models to become more robust against attacks. Finally, we will also try to study this security issue for applications of meta learning in other domains beyond the few-shot classification such fast reinforcement learning~\cite{finn2017one} and meta machine translation~\cite{gu2018meta}. 
\bibliographystyle{unsrt}
\bibliography{sample}
\end{document}